\pdfoutput=1

\documentclass[11pt]{article}

\usepackage{acl}

\usepackage{times}
\usepackage{latexsym}
\usepackage{booktabs}
\usepackage{graphicx}
\usepackage{enumitem}
\usepackage{multirow}
\usepackage{todonotes}
\usepackage{subfigure}
\usepackage{float} 
\usepackage[T1]{fontenc}

\usepackage[utf8]{inputenc}
\usepackage{amsmath}
\usepackage{siunitx}

\usepackage{microtype}
\usepackage[capitalise]{cleveref}

\title{The Curious Case of Control}
\author{Elias Stengel-Eskin \\ Johns Hopkins University \\ {\tt{elias@jhu.edu}} \And Benjamin Van Durme \\
Johns Hopkins University \\
{\tt{vandurme@jhu.edu}}}

\begin{document}
\maketitle
\begin{abstract}
Children acquiring English make systematic errors on subject control sentences even after they have reached near-adult competence \citep{chomsky.c.1969}, possibly due to heuristics based on semantic roles \citep{maratsos.m.1974}.
Given the advanced fluency of large generative language models, 
we ask whether model outputs are consistent with these heuristics, and to what degree different models are consistent with each other. 
We find that models can be categorized by behavior into three separate groups, with broad differences between the groups. 
The outputs of models in the largest group are consistent with positional heuristics that succeed on subject control but fail on object control.
This result is surprising, given that object control is orders of magnitude more frequent in the text data used to train such models. 
We examine to what degree the models are sensitive to prompting with agent-patient information, finding that raising the salience of agent and patient relations results in significant changes in the outputs of most models. 
Based on this observation, we leverage an existing dataset of semantic proto-role annotations \citep{white.a.2020} to explore the connections between  control and   labeling event participants with properties typically associated with agents and patients.\footnote{Code and prompts available at {\tt{\url{https://github.com/esteng/curious-case-of-control}}}.} 
\end{abstract}  

\section{Introduction} \label{sec:intro}
Normally-developing children learning English struggle with subject control clauses long after they have successfully acquired the components to understand them \citep{chomsky.c.1969, cromer.r.1970, maratsos.m.1974, sherman.j.1993}.
A sentence with a subject control clause has a matrix (main) clause containing a main verb, an agent, and a patient, and an embedded infinitival clause. 
For example, in \emph{Cole promised Joe to call},
the agent is \emph{Cole}, the patient is \emph{Joe}, and the embedded clause is \emph{to call}. 
Crucially, the embedded verb here does not have an overt subject, but rather implicitly refers to a subject in the matrix clause (in this case, \emph{Cole}).
In a subject control clause, this latent subject of the embedded infinitival clause (usually written as \emph{PRO}) is coindexed with the \emph{subject} (\emph{Cole}) rather than the object (\emph{Joe}) of the matrix (main) clause, i.e. \emph{Cole} (the agent) is doing the calling.  
This is typically written:
\vspace{-0.5em}
\begin{equation}
    \textstyle \text{[Cole]$_{\text{NP}_i}$ \textbf{promised} [Joe]$_{\text{NP}_j}$ PRO$_i$ to call} \label{eqn:subject}
    \vspace{-0.5em} 
\end{equation}
where subscripts indicate the noun phrase (NP) \emph{``Cole''} is the subject of \emph{``to call''}. 
(\ref{eqn:subject}) can be contrasted with the more common case of object control; for example, if the matrix verb \emph{``promised''} is swapped with an object control verb like \emph{``told''}, then the coreferrent of PRO changes: 
\vspace{-0.5em} 
\begin{equation}
    \text{[Cole]$_{\text{NP}_i}$ \textbf{told} [Joe]$_{\text{NP}_j}$ PRO$_j$ to call} \label{eqn:object}
    \vspace{-0.5em} 
\end{equation}
\citet{chomsky.c.1969}\footnote{NB: the author is Carol Chomsky, not Noam Chomsky.} finds that children ages 5 to 10 regularly misinterpret subject control (\ref{eqn:subject}) for object control (\ref{eqn:object}) while correctly interpreting object control clauses, and proposes that children are following the Minimal Distance Principal (MDP), choosing the linearly closest noun phrase (NP) to govern PRO.
\citet{cromer.r.1970} highlights the systematicity with which children mistake subject control for object control and provides evidence for the MDP. 
However, \citet{maratsos.m.1974} argues against the MDP; while his results support the observation that children struggle with subject control, they do not support the MDP, favoring an alternative based on semantic roles. 
\citeauthor{maratsos.m.1974} changes the subject and object order through passivization:
\vspace{-0.5em} 
\begin{equation}
 \text{[Joe]$_{\text{NP}_j}$ \textbf{was told} by [Cole]${_{\text{NP}_i}}$ PRO$_j$ to call} 
 \vspace{-0.5em} \label{eqn:passive} 
\end{equation}
finding that children \emph{correctly} coindex PRO with the (further away) object, violating the MDP.  

\begin{figure}
    \centering
    \includegraphics[width=\columnwidth]{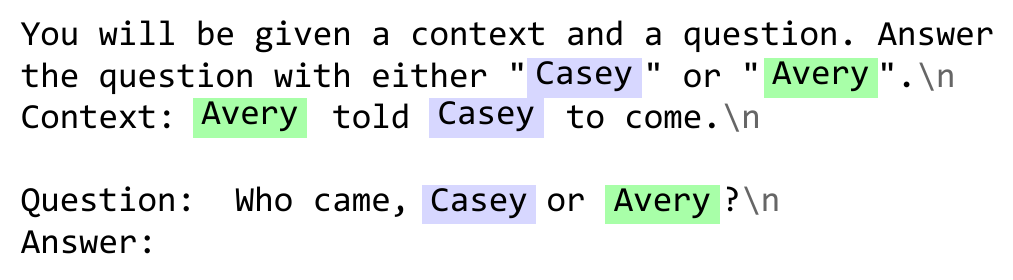}
    \vspace{-2em}
    \caption{Zero-shot probe for object control. Colors indicate names, which can be swapped.}
    \vspace{-1em}
    \label{fig:zero_shot_prompt}
\end{figure}

Recently, large pre-trained language models have shown an impressive ability not only to produce fluent text, but also to perform tasks by ``filling in the blank'' in question-answering prompts, either with no previous examples (zero-shot) or with a few representative examples (few-shot) \cite{brown.t.2020, raffel.c.2020, sanh.v.2021}. 
In light of the difficulty children have in acquiring subject control constructions, we explore how the outputs of the language models tested compare with adult and child strategies for coindexing PRO. 
In \cref{sec:exp1}, we examine this question in the zero-shot setting where we give the models only a single question, treating each model as a sort of experimental subject (cf. \cref{fig:zero_shot_prompt}).  
Our initial hypothesis is that model outputs will be consistent with child strategies, i.e. the models will perform well on object control examples, but misinterpret subject control for object control.
This is informed by two factors: object control is orders of magnitude more frequent than subject control (cf. \cref{sec:freq}), and active object control (i.e. (\ref{eqn:object})) requires resolving a shorter dependency than subject control. 
We instead find that the tested models fall into three groups, with the majority in fact producing responses mistaking object control for subject control -- the opposite of what children do. 

\begin{figure}[H]
    \centering
    \vspace{-1em}
    \includegraphics[width=\columnwidth]{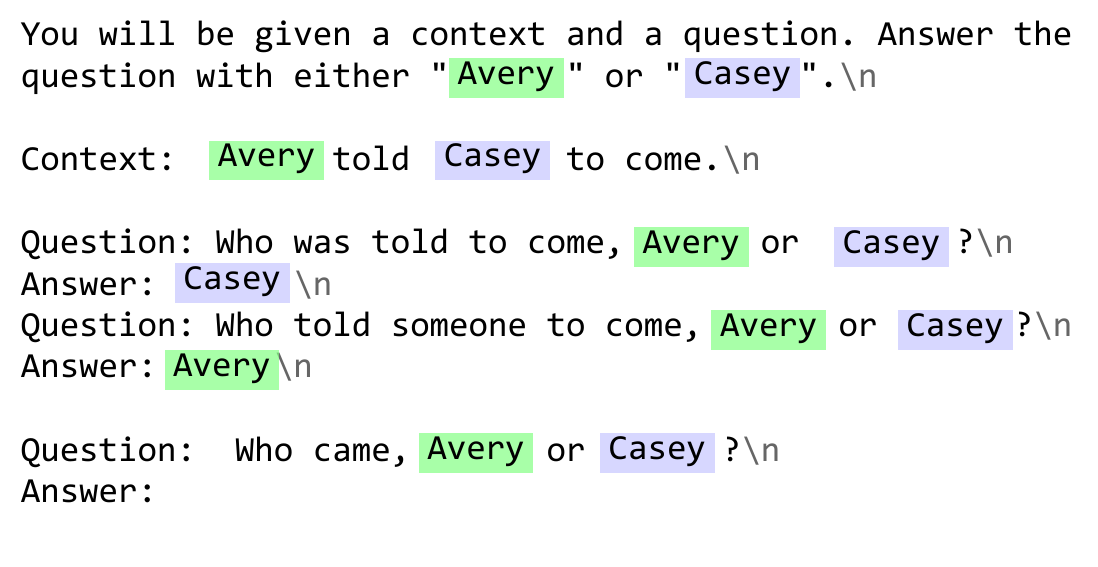}
    \vspace{-3em}
    \caption{A prompt-hacked example for object control, with long-form instructions.}
    \label{fig:hacked_prompt}
    \vspace{-1em}
\end{figure}

Following these observations, in \cref{sec:exp2} we examine to what degree this behavior is sensitive to semantic roles, following \citet{maratsos.m.1974}. 
To test this, we investigate a ``few-shot'' setting, where we prompt the model not only with the context and a single question, but also with a set of question-answer pairs that raise the salience of the matrix agent and patient (cf. \cref{fig:hacked_prompt}).
We find that the models whose behavior in the zero-shot setting was consistent with a positional heuristic have significant differences in the few-shot setting, and the directions of these difference are consistent with a sensitivity to semantic roles. 

Finally, in \cref{sec:exp_srl} we investigate whether the sensitivity to semantic roles corresponds to performance on a semantic proto-role labeling task, where models are tested with questions about volition and change of state, properties associated with agents and patients (respectively). 
We find that while some models are able to perform the labeling task surprisingly well, the differences between models do not necessarily map to the differences in \cref{sec:exp2}. We offer three key takeaways: 
\begin{enumerate}[noitemsep, nolistsep, topsep=0pt, leftmargin=*]
    \item For many models (all GPT-Neo variants, Jurassic Jumbo) the outputs are surprising given their training distributions; while object control is orders of magnitude more common in the text data used in training these models, they perform better on subject control.
    \item Large pretrained models are not consistent among themselves. 
    Even models with similar architectures can have very different trends in their outputs, and the outputs of autoregressive and text-to-text models differ substantially. 
    \item The associations in the autoregressive models tested form outputs that can be explained by simple, often position-based heuristics. 
    For text-to-text models (e.g. T0, T5) the output patterns can also be captured by heuristics based on agent and patient relations.
    However, sensitivity to agent and patient relations in subject and object control clauses does not always entail higher performance on semantic proto-role labeling. 
\end{enumerate}

\section{Models and Metrics} \label{models}
We explore both autoregressive models and text-to-text models.  
Autoregressive models are optimized by minimizing $-\log(P(w_i | w_{-i}))$ for words $w_1, \ldots w_N$ in a given context. 
These models (also called ``decoder-only'' models) are composed of just a decoder, which encodes the previously observed tokens $w_1, \ldots w_{i-1}$ produces a probability distribution over the vocabulary for the next token, $w_i$.
Text-to-text models are encoder-decoder models, optimized to reconstruct a noised version of the input via the decoder.
An encoder takes a corrupted version of whole sequence $w_1, \ldots w_N$ as input, encoding it into a dense representation from which the decoder reconstructs the original $w_1, \ldots w_N$.

The autoregressive models considered are:
\begin{itemize}[noitemsep, nolistsep, topsep=0pt, leftmargin=*]
    \item \textbf{GPT-3 Davinci}: this model is only available through the OpenAI API, and its exact training details are unclear. 
    It is based on the GPT-3 model \citep{brown.t.2020} which was trained on Common Crawl \citep{raffel.c.2020} with 175B (billion) parameters. 
    Among the several versions of GPT-3, Davinci is generally regarded as the highest-performing \citep{openai}. 
    \item \textbf{GPT-Neo}: this is an open-source replication of GPT-3 introduced by \citet{black.s.2021}, trained on The Pile \citep{gao.l.2020}, a 800Gb dataset of web-text intended for pre-training. 
    GPT-Neo has 3 sizes: 1.3B, 2.7B, and 6B parameters (GPT-J), all trained on the same dataset, allowing for direct comparison. 
    \item \textbf{Jurassic}: Jurassic Large (7.5B parameters) and Jurassic Jumbo (178B parameters) \citep{opher.l.2021} are also accessible only through an API. 
    The training data is based on Common Crawl, though similarly to GPT-3 Davinci, the details of the training data filtering process are unclear.
    Relevant differences to GPT-3 are in the tokenization (which includes multi-word expressions) and use of fewer, wider layers. 
\end{itemize}

The text-to-text models we consider are:
\begin{itemize}[noitemsep, nolistsep, topsep=0pt, leftmargin=*]
    \item \textbf{T5 for QA}: The T5-base text-to-text model (220-million parameters) \citep{raffel.c.2020} was pre-trained on cleaned Common Crawl data (C4) and fine-tuned on SQuaD question answering data \citep{rajpurkar.p.2016}. 
    \item \textbf{T0pp}: presented by \citet{sanh.v.2021}, T0pp is an 11B parameter model with a T5-like architecture, pre-trained on Common Crawl data and finetuned specifically for zero-shot question answering on the P3 dataset of NLP benchmarks. 
    This dataset recasts a large number of NLP benchmark datasets into question answering prompts. 
\end{itemize}
Note that because of fine-tuned nature of the ``T5 for QA'' model, the expected prompt format is fixed, unlike the unrestricted prompt format for the other models. 
Thus, prompt hacking cannot be done on this version of T5, and so it is only used in \cref{sec:exp1}. 
We access non-API models via the Transformers library \citep{wolf.t.2020}; due to computational constraints, they are run on single GPUs at 1/2 precision. 
For all models, we decode with the temperature parameter set to 0. 

\vspace{-0.5em}
\subsection{Metrics} \label{metrics}
\vspace{-0.5em}
Online APIs make forced decoding very costly \citep{shin.r.2021}. 
Rather than comparing logits for a restricted output vocab, we allow the model to freely generate tokens, letting the model  produce a larger variety of answers.
In other words, rather than comparing the output probabilities for particular tokens (the logits) given a fixed prefix, we compare full strings of output tokens. 
However, this method requires heuristics to classify the output strings into categories.
In \cref{sec:logits} we validate our heuristics, verifying that for locally-run models the trends are similar when using logits. 

Our metric first extracts single word answers and then searches for answers like ``The answer is: NAME''. 
For some models and settings, the model re-generates the entire prompt before answering, i.e. it copies the instructions, context, and question, before copying the answer continuation and finally producing an answer. 
We use Levenshtein distance to check whether the prompt has been regenerated; if it has, it is removed and the first string following the prompt is checked for answer strings. 
The extraction function always returns the first valid answer produced by the model.
If the extraction function fails to find any valid answer strings, the example is skipped in evaluation rather than counted as wrong.  
We measure significance in model differences with McNemar's test \citep{mcnemar.q.1947}, following \citet{dietterich.t.1998}. 

\section{Experiment 1}
In order to examine what types of generalizations are made by the examined models when prompted for subject and object control information, we construct a number of question-answering-style prompts, where the models fill in the answer. 
This approach follows recent literature \citep{raffel.c.2020, brown.t.2020} and takes advantage of the models question-answering abilities.
Moreover, using models pretrained with a language-modeling loss rather than training a model specifically for control lets us examine what types of generalizations are captured by the models' associations learned from its original training data, rather than whether a very large model can learn to answer subject and object control questions correctly.
We first describe the construction of the prompts used in this set of experiments.
Using those prompts, we validate the choice to use heuristic extraction from open generation (i.e. allowing the model to generate tokens up to an end-of-sequence token) rather than logits and confirm that in the C4 dataset, object control is more frequent than subject control. 
Then, we analyze the zero-shot performance of the models on the subject and object control prompts. 

\vspace{-0.5em}
\subsection{Subject and Object Prompts} \label{sec:methods}  
\vspace{-0.5em}
While pretrained language models used for QA are often evaluated in a ``few shot'' setting, where they are given a few ``training'' prompts  before answering a ``test'' prompt, in our main experiments we focus on the zero-shot setting.
This is in order to avoid learning effects that might result from few-shot prompting (as one would with human subjects) and follows human experiment paradigms, where experimenters are careful not to provide feedback to subjects about the expected answer until after all trials are complete. 
The prompts used in \cref{sec:exp1} and \cref{sec:exp2} have an instruction sentence, a context (like (\ref{eqn:subject})-(\ref{eqn:passive})), a question (e.g. ``Who called?''), and an answer continuation.
See \cref{fig:zero_shot_prompt} for an example zero-shot object control prompt. 

We take the max over two instruction types (long and short) in our analyses. \cref{fig:zero_shot_prompt} shows the long-form instructions, which include the options in the instructions (e.g. \emph{Answer the question with either "Casey" or "Avery"}) and in the question (e.g. \emph{Casey or Avery}). 
The short-form instructions omit these prompts in the instructions and questions. 

Since the models examined can be sensitive to specific tokens, we cover 9 embedding verbs for object control, chosen from a selection of common linguistics examples: ``told'', ``ordered'', ``called upon'', ``urged'', ``asked'', ``persuaded'', ``convinced'', ``forced'', and ``pushed''. 
These verbs are presented both in the active (object control experiments) and passive (passive object control experiments). 
These include verbs that trigger a factuality inference in the affirmative (e.g. ``persuaded'', ``convinced'', ``forced''). 
For subject control, we follow previous work \citep{chomsky.c.1969, maratsos.m.1974} and use ``promise''; we also include ``threaten'' as a subject control verb.
These verbs are presented only in the active voice, as sentences such as ``Casey was promised by Avery to call'' were deemed too ambiguous (if grammatical at all). 
In our main experiments, we use names as NPs; we also report results in \cref{append:profession} using common professions to ensure that the trends observed with names hold. 
We chose 2 male names, 2 female names, and 2 gender-neutral names; these were chosen by taking the top 2 names in each reported gender category in US Social Security data from 1970 to 2019.\footnote{\url{https://www.ssa.gov/oact/babynames/}}
The gender-neutral names were chosen by taking the top names in the intersection of male and female names.\footnote{We note that this does not guarantee that the name is equally likely for both genders.}
We run the same prompt with each name combination in both orders, to avoid possible biases the model may have towards particular names. 
When the names are included in the instruction, we add an example with the name order swapped to avoid confounding due the model simply copying the first or last name to appear in the instructions. 
Finally, for the action infinitive (i.e. the embedded verb) we chose the first 5 coherent verbs (i.e. intransitive infinitives) from a frequency list of English verbs \citep{yu.c.2020, sharov.s.2020}.
This yields 1500 sentences for object control and 150 for subject control (3000 and 300 with swapped names). 

\subsection{Results and Analysis}\label{sec:exp1} 
\paragraph{Frequency of Subject and Object Control} \label{sec:freq} 
In \cref{sec:intro}, we claimed that object control is more frequent that subject control. 
To support this claim in the context of the models examined here, we conduct a search of a subset of the C4 dataset \citep{raffel.c.2020} for sentences fitting subject control and object control templates. 
While there are many types of subject and object control, we focus on infinitival complements with transitive matrix verbs, searching with templates similar to the sentences in examples \ref{eqn:object} and \ref{eqn:subject}. 
For object control, we use the same verb list as in \cref{sec:methods}. 
For subject control, we use ``promise'' and ``threaten'', as in \cref{sec:methods}. 
We allow the embedded verb to be any verb. 
We sub-sample the first \num{1000000} sentences of C4 and search them with the templates, finding that object control occurs \num{10435} times, while subject control occurs only \num{209} times, i.e. object control is $\sim 50$ times more frequent.\footnote{Note that object control is to be distinguished from Exceptional Case Marking \citep[ECM; ][]{chomsky.n.1986} and raising, which yield similar surface forms but have a different analysis than object control. 
Since object control is already shown to be far more frequent than subject control, we do not examine common ECM and raising verbs; however, we note that, when used with transitive matrix verbs (i.e. verbs with a matrix subject and object) these constructions generally follow the object control template, where the matrix object is the embedded subject.}  

\paragraph{Validating Logits} \label{sec:logits}

\begin{figure}
    \centering
    \includegraphics[width=\columnwidth]{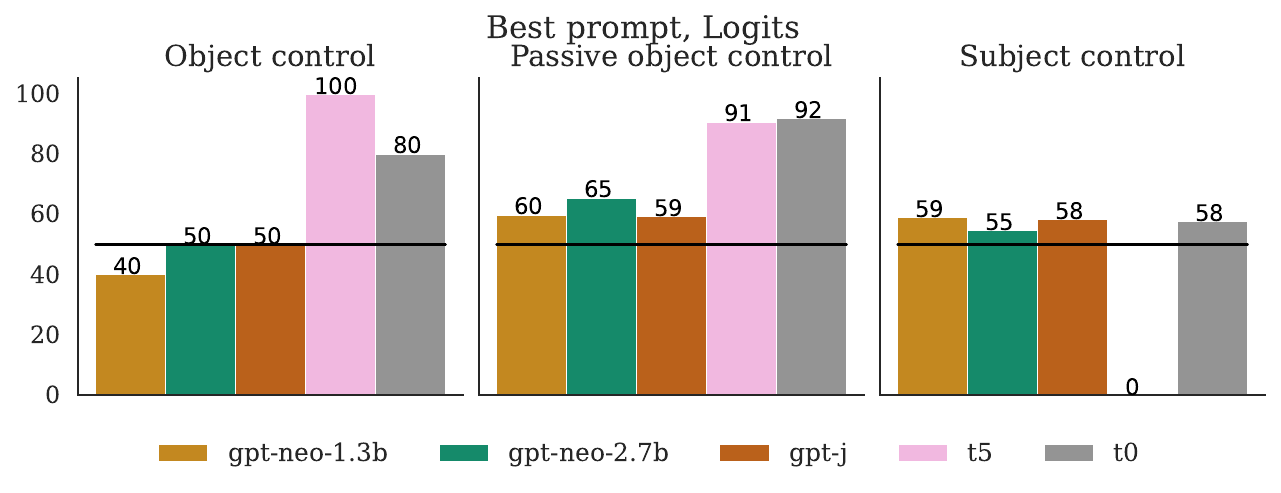}
    \vspace{-2em}
    \caption{Accuracy of logit-scored model, taking the max across instruction types.}
    \label{fig:logits}
    \vspace{-1em}
\end{figure}

\cref{fig:logits} shows the zero-shot accuracy of the best instructions using logit scoring, for the models for which we have access to the full output distribution (non-API models). 

\begin{figure}[ht]
    \centering
    \includegraphics[width=0.5\textwidth]{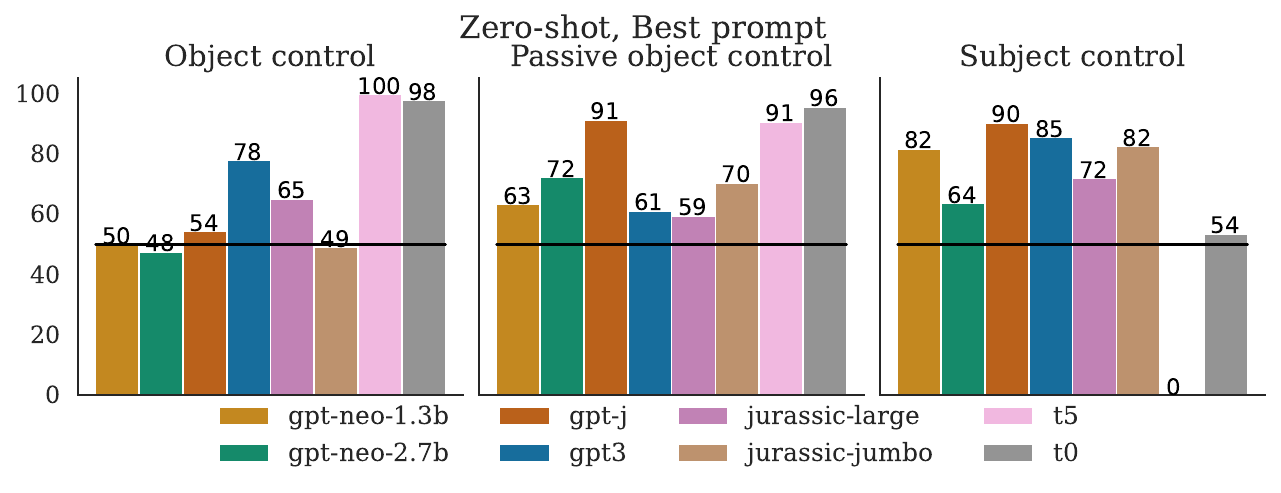}
    \vspace{-2em}
    \caption{Zero-shot accuracy on object control, passive object control, and subject control. Black line represents random performance ($50\%$ accuracy).}
    \label{fig:zero_shot_results}
    \vspace{-1em}
\end{figure}

Comparing the results to \cref{fig:zero_shot_results}, we see similar but less pronounced trends.
As in \cref{fig:zero_shot_results}, GPT-Neo models perform better on subject control and passive object control, while text-to-text models perform better on object control and passive object control. 
This validates our choice to use heuristics in later experiments. 
We note also that for GPT-Neo models, the heuristic results in \cref{fig:zero_shot_results} reflect higher accuracies than the logit-based results in \cref{fig:logits}, indicating that the heuristics capture broader range of outputs corresponding to valid answers. 

\vspace{-0.5em}
\paragraph{Zero-shot performance}
In \cref{fig:zero_shot_results}, we see that model classes have different results; we further classify models into 3 groups:
\begin{enumerate}[noitemsep, nolistsep, topsep=0pt, leftmargin=*]
\item The GPT-Neo variants and Jurassic Jumbo are better on subject and passive object control than object control. This pattern can be accounted for by a positional heuristic, namely to take the \emph{first} NP in the matrix clause (i.e. \emph{Maximum} Distance Principle rather than the minimum distance principle of \citet{chomsky.c.1969}). 
\item T5 and T0 are consistent with the observations in \citet{maratsos.m.1974}; both models do better on object control (active and passive) than subject control. This contradicts the MDP but is consistent with a heuristic choosing the matrix patient. 

\item GPT-3 and Jurassic Large both perform well above chance on object control and subject control, with their best performance on subject control, but both perform worse on passive object control. This could be matched to a positional heuristic (take the second NP) for object control verbs, rather than an agency-based heuristic. 
\end{enumerate}

\vspace{-0.5em}
\paragraph{Further observations} 
Even within model families, there are measurable differences: although GPT-3 and Jurassic Jumbo are roughly the same size and share a general architecture, and are ostensibly trained on similar data, the changes made by \citet{opher.l.2021} seem to have a measurable impact, with Jurassic Jumbo performing differently on zero-shot object control examples. 
For active object control, the difference $\Delta_{\text{GPT-3}, \text{Jumbo}}^{\text{active}} = acc_{\text{GPT-3}} - acc_{\text{J-Jumbo}} = 29$ ($p < 0.01$), and for passive $\Delta_{\text{GPT-3}, \text{Jumbo}}^{\text{passive}} = -9$ ($p < 0.01$).
Similarly, GPT-3 differs from GPT-Neo-1.3B on active object control, and from GPT-Neo-2.7B and GPT-J on both forms of object control, despite sharing an architecture. 
Further analysis is impeded by a lack of details on the data used to train GPT-3 and Jurassic; this underscores the need for model creators to be transparent about training data and details. 

\begin{figure}[ht]
    \centering
    \includegraphics[width=0.5\textwidth]{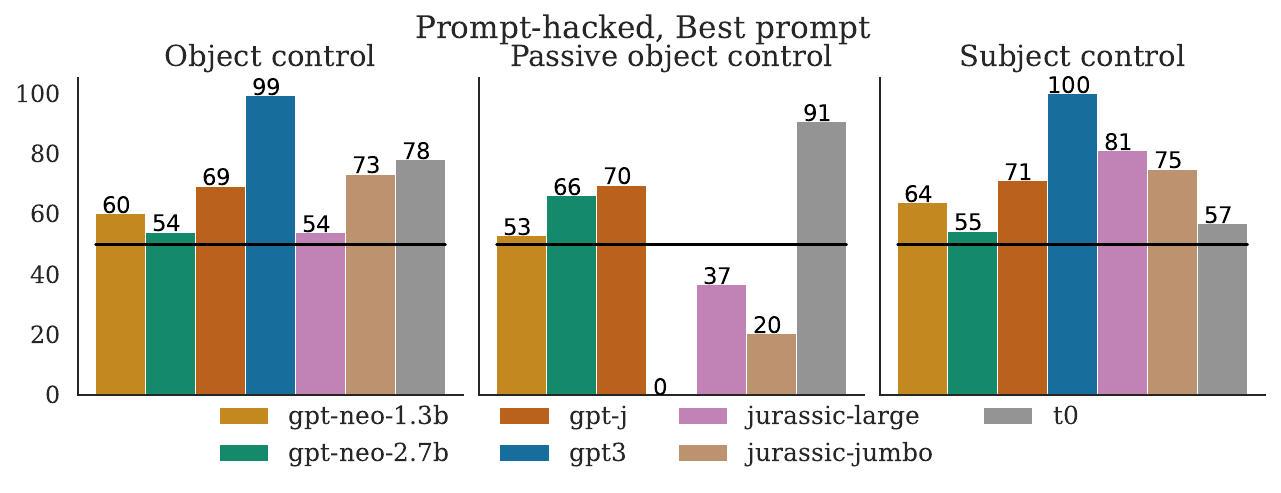}
    \vspace{-2em}
    \caption{Accuracy on object control, passive object control, and subject control after prompting with agent and patient questions. Accuracy changes from \cref{fig:zero_shot_results} are generally consistent within heuristic groups.}
    \label{fig:hacked_results}
    \vspace{-0.75em}
\end{figure}

We also observe that larger models tend to have higher performance: GPT-J is better on all settings than GPT-Neo-1.3B and 2.7B, and Jurassic Jumbo is better than Jurassic Large on passive object control. 
That said, some larger models are also slightly worse than their smaller counterparts (e.g. Jurassic Jumbo on object control). 
This suggests that larger models may be more prone to learning patterns corresponding to simple heuristics; however, additional evidence is needed. 

\section{Experiment 2} \label{sec:exp2} 
Recent work has shown that providing examples in the prompt to a frozen pre-trained model can yield higher performance on QA tasks \citep{brown.t.2020}.
In this setting, typically called ``in context learning'', some number of demonstrations of questions and answers are given in the context, followed by a test example, to which the model produces an answer.
The demonstrations in the context give the model additional guidance on which task is being evaluated. 
While we do not do in-context learning with training questions about object in subject control, we do experiment with adding information to the prompt to raise the salience of agents and patients (e.g. ``Q: Who told someone to call? A: Cole'' for (\ref{eqn:object}).) 
An example can be seen in \cref{fig:hacked_prompt}; the QA pairs given in the context give the agent and the patient of the matrix clause, but not the embedded clause.
This can be thought of as a form of chain-of-thought prompting \citep{wei.j.2022} or a scratchpad \citep{nye.m.2021} where the gold answers are provided.

We hypothesize augmenting the prompt with agent-patient questions will affect each group as follows:
\begin{enumerate}[noitemsep, nolistsep, topsep=0pt, leftmargin=*]
\item For Group 1 (GPT-Neo, Jurassic Jumbo) where the models' outputs are consistent with positional heuristics, the additional prompts will provide some evidence inconsistent with the heuristic. This evidence may lead to changes in the models' associations that result in outputs less consistent with the heuristics. 
For example, in \cref{fig:hacked_prompt} the question \emph{Who was told to come, Avery or Casey?} the answer \emph{Casey} provides evidence against taking the furthest-away NP as the answer.
In that case, we would expect a drop in performance in passive object control and in subject control, where the heuristic is beneficial, and an increase in object control, where the heuristic does not help.

\item In Group 2 (T0 and T5), since the model outputs are already consistent with a semantic role-based explanation, we do not expect much change in any setting. 
In other words, the model outputs are already consistent with access to agent-patient relations combined with an incorrect association for outputting the matrix clause's agent as the embedded subject. 

\item Finally, in Group 3 (GPT-3, Jurassic Large), we see that the models' outputs on object control are consistent with an object control-specific heuristic (to take the first NP) the models have lower performance for passive object control than active object control. 
Thus, as in Group 1, we expect that evidence against the positional heuristic in the prompts will boost performance in passive object control, while reducing it on active object control. 
\end{enumerate} 

\vspace{-0.5em}
\subsection{Results and Analysis} 
\vspace{-0.5em}
\cref{fig:hacked_results} shows the results after applying prompts with questions about agents and patients. 
Here, we see that for Group 1 (GPT-Neo and Jurassic Jumbo) the performance does decrease for subject control and passive object control. This decrease is significant for all models and settings ($p < 0.01$) except Jurassic Jumbo in subject control ($\Delta_{\text{Jumbo}}^{\text{subj}} = -7$, $p = 0.06$). 
At the same time, all object control performance increases significantly for Group 1 ($p < 0.01$). 
These results confirm our hypotheses, supporting the notion that these models are consistent with a positional rather than semantic heuristic. 
For Group 2 (T0), we find a significant decrease in performance on both object control types ($\Delta_{\text{T0}}^{\text{active}} = -20$, $p < 0.01$), and no significant difference for subject control ($p = 0.14$); this is roughly consistent with our predictions, since the size of the decrease for object control is relatively small (e.g. compared to the decrease for GPT-3 in the passive). 
Finally, for Group 3 (GPT-3 and Jurassic Large) we largely see the opposite of what we expected: GPT-3's performance on object control goes close to ceiling after prompting with agent-patient questions, while the passive performance drops to 0; similarly, Jurassic Large's performance drops on passive object control, dropping slightly also on active object control.
Both models improve on subject control, with significant differences from the zero-shot setting ($\Delta_{\text{GPT-3}}^{\text{subj}} = 15$, $p < 0.01$, $\Delta_{\text{Large}}^{\text{subj}} = 9$, $p < 0.01$), perhaps reflecting an effect of the semantic role-based priming. 
Note that for GPT-3, the passive performance drop is from a lack of parseable, non-empty strings being produced, rather than incorrect predictions. 
For active object control, GPT-3 outputs the second NP more often with additional prompts, increasing the score from $78\%$ to $99\%$.
It may be that the associations responsible for this are also to blame for the degenerate behavior seen in the passive, where the model produces only an end-of-sequence token. 

\section{Experiment 3} \label{sec:exp_sprl}

Following \citet{maratsos.m.1974}'s hypothesis that the observed mistakes children make on subject control sentences is driven by semantic roles, in \cref{sec:exp_srl} we examine the relationship between a model's ability to perform zero-shot object and subject control and its accuracy on identifying attributes commonly associated with agents and patients. 
Querying language models using fixed semantic role ontologies (e.g. AGENT, PATIENT, THEME) may be difficult as these ontologies may be absent from pretraining corpora.
We instead measure the models ability to perform semantic proto-roles labeling (SPRL) for the volition and change of state properties. 
We use the SPRL data provided in the Universal Decompositional Semantics (UDS) dataset introduced by \citet{white.a.2020}. 
These properties, first proposed by \citet{dowty.d.1991}, were found to be strongly prototypical of agents and patients, respectively \citep{reisinger.d.2015}.\footnote{While instigation and stationarity were slightly more predictive of agency and patienthood, they were deemed to be more difficult to re-frame as a prompt.} 
Proto-role inferences are elicited with simple prompts, circumventing brittle and complicated ontologies. 
Indeed, the UDS dataset was built by asking annotators questions like \emph{``How likely is it that ARG chose to be involved in the PRED?''} and normalizing their scalar ratings to $[-3, 3]$. 

\subsection{Semantic Proto-Role Labeling Prompts} 
To construct a dataset of SPRL prompts, 
we first filter the UDS dataset for sentences with $< 35$ tokens -- this eliminates many long sentences, which are often more difficult to answer.
We then eliminate examples with scalar annotations $\in (-1,1)$, keeping only examples with strong inferences about the properties.
The annotations are binarized with values $\geq 1$ leading to a ``Yes'' answer and values $\leq -1$ leading to ``No''.
The annotations are balanced between ``Yes'' and ``No'', with the excess examples from the more frequent category being removed.

\begin{figure}[ht]
    \centering
    \includegraphics[width=\columnwidth]{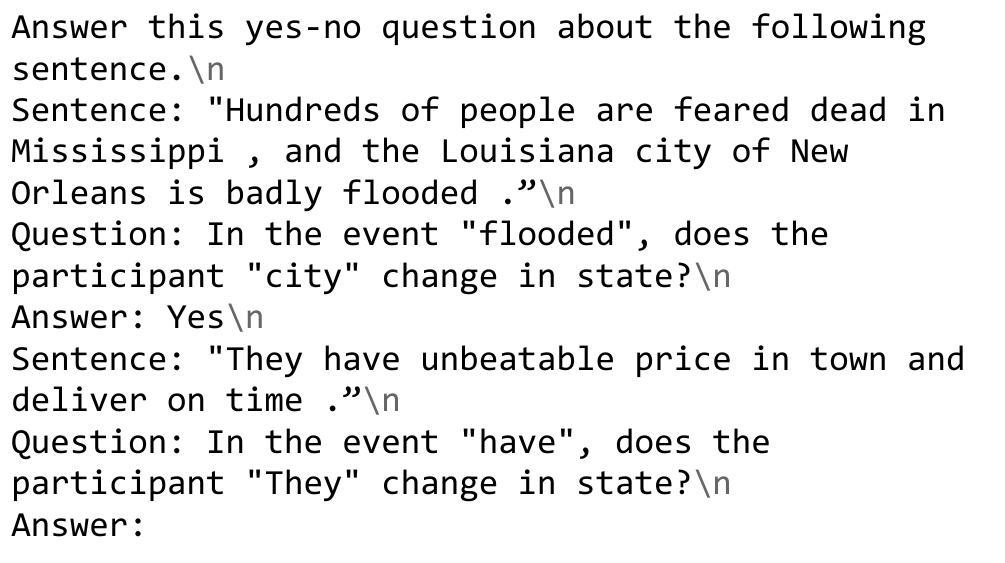}
    \vspace{-2em}
    \caption{Prompt for eliciting SPRL judgments, shown here with one prompting example (1-shot).}
    \label{fig:sprl_prompt}
    \vspace{-1em}
\end{figure}

Two templates are used for each property; an example template is shown in \cref{fig:sprl_prompt}.
For volition, one template asks, `\emph{In the event ``PRED'', does the participant ``ARG'' act with volition?}' while the other asks `\emph{In the event ``PRED'', does the participant ``ARG'' act on purpose?}'. 
For change of state, the first template asks, `\emph{In the event ``PRED'', does the state of the participant ``ARG'' change?}' and the other asks, `\emph{In the event ``PRED'', does the participant ``ARG'' change in state?}'
We take the maximum over these two templates. 

In our first set of experiments we are interested in the raw ability of the model to perform the semantic proto-role labeling (SPRL) task, and so we allow for full prompt hacking, where demonstrations of the task are provided as part of the context.
Accordingly, we stratify the annotations into 4 stages; the bottom stage always forms the ``test'' prompt, with the answer blank. The remaining 3 stages are added for increasing levels of in-context learning with ``training'' question-answer pairs
(i.e. when the 3rd layer is added, there are 3 example question-answer pairs with answers, and one ``test'' pair that has no answer, where the model must fill in the answer.) 
\cref{fig:sprl_prompt} shows a prompt with one training stage, followed by one test example. 
We ensure that we use each annotation only once, and that all test annotations paired across across prompting settings. This results in 118 change-of-state test prompts and 168 volition prompts. 

\paragraph{Hypotheses}
We expect that models which perform well on zero-shot subject and object control (e.g. those that can model both the active and passive of object control) will also have higher performance on SPRL, since both require semantic role information.
Specifically, we expect to see higher performance from T0 and T5 on at least one of the properties, since their outputs on active and passive object control are consistent with a heuristic that identifies patients as embedded subjects, rather than a positional heuristic. 
Thus, it may be that the representations learned by T0 and T5 contain more information on agency and patienthood, leading to better performance on SPRL. 

\subsection{Results and Analysis} \label{sec:exp_srl}
\cref{tab:sprl} shows the accuracy on binary semantic proto-role labeling of all models with performance significantly above a random baseline.
\begin{table}[h]
    \centering
    \begin{tabular}{ccccc}
    \toprule
   Setting & Model & $\#$ shots & Acc. & $\#$ valid\\
    \midrule
    $\Delta$State &  GPT-3 & 1 & \num{0.61} & \num{118} \\
    \midrule
    \midrule
   \multirow{3}{*}{Volition}  & GPT-3 & 3 & \num{0.77} & \num{168} \\
    & GPT-J & 0 & \num{0.69} & \num{111} \\
    & T0 & 0 & \num{0.60} & \num{168} \\
    \bottomrule
    \end{tabular}
    \vspace{-0.5em}
    \caption{Accuracy on change-of-state and volition for models significantly above random baseline.}
    \label{tab:sprl}
\end{table}
For change of state, only GPT-3 performs above chance, while for volition, GPT-3, GPT-J, and T0 perform above chance. 
T0's lower performance is surprising, as the performance of T0 in \cref{fig:zero_shot_results} is more consistent with an role-based heuristic. 
In other words, we do not find that models performing well on both passive and active control perform well on SPRL. 
However, these are separate tasks -- thus, it is possible for GPT-3 and GPT-J to be consistent with non-role-based heuristics in one task while still encoding information about agent and patient properties. 
Finally, we note that in both \cref{fig:zero_shot_results} and \cref{fig:hacked_results}, GPT-3 performs well on subject control and object control in the active, which is consistent with it containing information on agency and patienthood. 

\section{Related Work} 
Following the advent of pretrained language models there has been an explosion of work examining what kinds of linguistic knowledge such models contain.
\citet{rogers.a.2020} provide a comprehensive survey of probing work on BERT, covering probing for syntactic, semantic, and world knowledge.  
This line of probing work generally makes use of linear classifiers on top of frozen representations, tuned on a training set \citep{adi.y.2016, hupkes.d.2018, hewitt.j.2019}.

In contrast, we follow more recent work in probing large generative models using cloze-style prompts and relatively open generation \citep{schick.t.2021}.  
Such models (generative and non-generative) have been probed for diverse knowledge, including syntax \citep{futrell.r.2019}, symbolic reasoning \citep{talmor.a.2020}, and common-sense knowledge \citep{petroni.f.2019, kassner.n.2020, sakaguchi.k.2020}.
This has often been done by recasting benchmark datasets into text, either with zero examples \citep{sanh.v.2021} or in the form of in-context learning \citep{brown.t.2020, raffel.c.2020}.
\citet{ettinger.a.2020} present a suite of comparisons between pretrained language models and psycholinguistic experiments.
In a similar vein to our work, \citet{lee.s.2022} examine GPT-2's performance on reflexive anaphor agreement in subject and object control clauses, finding that GPT-2 performs well on object control but not transitive subject control; we do not examine reflexive anaphora, and expand our analysis to multiple model classes.  

Language models can be sensitive to the format of a prompt -- in order to improve extraction of relational knowledge from large language models, \citet{jiang.z.2020} propose automatic methods for mining new prompts and paraphrasing existing prompts. 
Similarly, \citet{qin.g.2021} propose a method for gradient-based prompt optimization, and \citet{shin.t.2020} propose a gradient-based search for prompt token replacement.
As our experiments require a specific prompt syntax, we choose to instead run prompts across different instruction styles and name-verb combinations.
Large language models are also sensitive to the frequency of terms in their training data; \citet{razeghi.y.2022} show a strong correlation between frequency of a term in a corpus and performance on tasks requiring that term. 
This further hightlights how surprising it is that several models perform better on subject control than object control.

\section{Conclusion} 
The results in \cref{fig:zero_shot_results} indicate that differences between models are not merely of degree, but of kind, with groups of models following different patterns, many of which are inconsistent with the dominance of object control in English. 
This highlights the pressing need for transparency in the reporting of model details, and especially of training data, without which it is impossible to hypothesize \emph{why} these differences are observed. 
We also find that, despite there being no trainable parameters in the few-shot setup of \cref{fig:hacked_results}, the models tested are in many cases predictably sensitive to semantic role information.
Our results in \cref{sec:exp_sprl} suggest that some models appear able to perform semantic proto-role labeling for volition and change of state (\cref{tab:sprl}), but this ability is not directly tied to the sensitivity to semantic roles in \cref{fig:hacked_results}.
In other words, some models contain information on semantic roles but may not recruit that information in producing answers for control examples. 
This leads to an interesting direction of future work in applying causal mediation analysis \citep{vig.j.2020,  elazar.y.2021} to control clauses, to disentangle the information present in a model from the process by which the model produces an output. 

\section{Limitations}
Firstly, this work is limited by its focus on English syntax, models, and examples. 
Control constructions exist a variety of languages \citep{landau.i.2001}; unfortunately, large pre-trained models currently exist primarily in English alone.
Another limitation is the use of fixed prompts: all models tested were found to be sensitive to the prompt format, and while a large number of prompts were explored by varying instructions, names, verbs, and actions, there may be more optimal prompts for the task. 
Our work is also limited by the use of open generation. While open-ended generation allows for more flexibility than constrained decoding, it introduces the challenge of interpreting the model outputs, though we do validate the use of open generation in \cref{sec:logits}.
We note that both these limitations are also common in human subject research.

While our results show that model outputs are consistent with simpler heuristics, some of which are observed in human children, we have attempted to clearly separate this from any anthropomorphic claims that the models might be actively ``following'' such a heuristic. 
The claim made is that the associations learned from large-scale pretraining  lead to outputs with patterns that can be described by simple heuristics, and that those heuristics at times differ from or resemble heuristics seen in human data. 
We also note that while we do not make strong commitments to any particular account of human language acquisition, all such accounts differ substantially from how the models tested are trained, i.e. on extremely large text-only corpora. 

\section*{Acknowledgements} 
Elias Stengel-Eskin is supported by an NSF Graduate Research Fellowship. 
This work was supported by NSF $\#1749025$.
We would like to thank Madelaine Thomas and Kate Sanders for feedback on previous drafts, as well as the ARR reviewers for their constructive comments and suggestions. 

\bibliography{control}
\bibliographystyle{acl_natbib}

\newpage  $\:$ \newpage
\appendix

\section{Additional Controls} \label{append:profession}
To ensure that our results are not biased by the use of names, we replicated a round of experiments using common professions rather than names. 
The professions used were ``doctor", ``lawyer", ``engineer", ``writer", ``janitor", ``bartender", e.g. \emph{``The bartender told the engineer to come.''}

\begin{figure}[ht]
    \centering
    \includegraphics[width=\columnwidth]{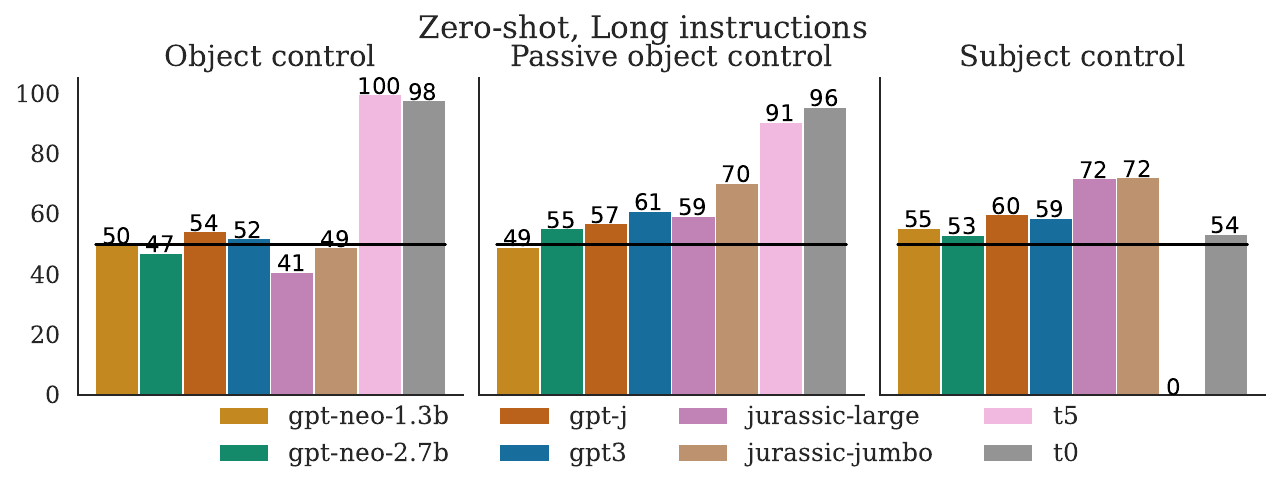}
    \caption{Accuracy of long instruction template on names, for comparison to \cref{fig:results_profession}}
    \label{fig:just_long}
\end{figure}

\begin{figure}[ht]
    \centering
    \includegraphics[width=\columnwidth]{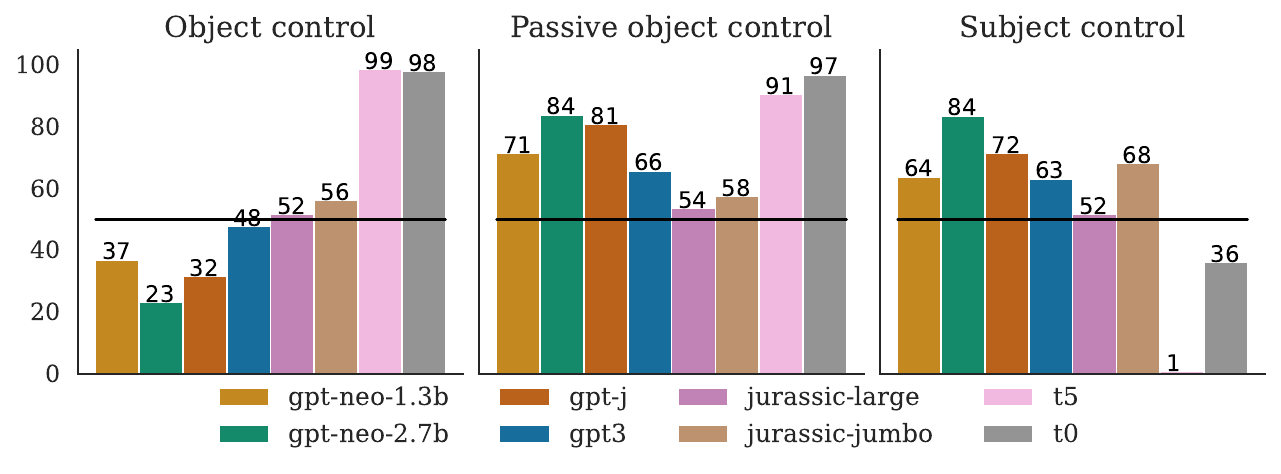}
    \caption{Accuracy of long instruction template on professions. Performance follows similar trends to comparable results with names (\cref{fig:just_long}). }
    \label{fig:results_profession}
\end{figure}

Profession prompts were run only with the long instruction format. 
In \cref{fig:just_long} we include for reference the results from the name experiments shown in \cref{fig:zero_shot_results} but restricted to only the long prompt format (\cref{fig:zero_shot_results} took the best accuracy over short and long instructions). 
\cref{fig:results_profession} shows that similar trends can be seen when using professions rather than names, confirming that the results are not due to name-specific processing.

\section{Licensing}
All data and code is released under an MIT license.

\end{document}